\documentclass[sigconf]{acmart}

\acmVolume{0}
\acmNumber{0}
\acmArticle{0}
\acmMonth{0}

\acmConference[CHASE'25]{Make email}{Jun 24 2025 - Jun 26 2025}{Manhattan, New York}

\usepackage{lipsum}
\usepackage{lscape}
\usepackage{xcolor}
\usepackage{algorithm}
\usepackage{algorithmicx}
\usepackage{amsmath}
\usepackage{hyperref}
\usepackage{hyperxmp}
\usepackage{titlesec}  
\usepackage{enumitem} 

\definecolor{linkblue}{RGB}{0,112,193}

\AtBeginDocument{%
  }

\titlespacing{\section}{0pt}{*1.0}{*0.8}   
\titlespacing{\subsection}{0pt}{*0.8}{*0.6} 
\titlespacing{\subsubsection}{0pt}{*0.6}{*0.4} 

\makeatletter
\let\@authorsaddresses\@empty
\makeatother

\begin{document}


\newcommand{\model}{PulseRide} 
\newcommand{\shortname}{\model{}: A Robotic Wheelchair for Personalized Exertion Control}

\title[\shortname]{\model{}: A Robotic Wheelchair for Personalized Exertion Control with Human-in-the-Loop Reinforcement Learning}


\author{Azizul Zahid}
\email{azahid@vols.utk.edu}
\orcid{0009-0008-0981-5798}
\affiliation{%
  \institution{University of Tennessee Knoxville}
  \streetaddress{P.O. Box 1212}
  \city{Knoxville}
  \state{Tennessee}
  \country{USA}
  \postcode{37921}
}

\author{Bibek Poudel}
\email{bpoudel3@vols.utk.edu}
\orcid{0000-0003-1535-0743}
\affiliation{%
  \institution{University of Tennessee Knoxville}
  \streetaddress{P.O. Box 1212}
  \city{Knoxville}
  \state{Tennessee}
  \country{USA}
  \postcode{37921}
}

\author{Danny Scott}
\email{dscott57@vols.utk.edu}
\orcid{0000-0002-5042-283X}
\affiliation{%
  \institution{University of Tennessee Knoxville}
  \streetaddress{P.O. Box 1212}
  \city{Knoxville}
  \state{Tennessee}
  \country{USA}
  \postcode{37921}
}

\author{Jason Scott}
\email{jlscott@utk.edu}
\orcid{0000-0001-7532-7073}
\affiliation{%
  \institution{University of Tennessee Knoxville}
  \streetaddress{P.O. Box 1212}
  \city{Knoxville}
  \state{Tennessee}
  \country{USA}
  \postcode{37921}
}

\author{Scott Crouter}
\email{scrouter@tennessee.edu}
\orcid{0000-0003-1297-9538}
\affiliation{%
  \institution{University of Tennessee Knoxville}
  \streetaddress{P.O. Box 1212}
  \city{Knoxville}
  \state{Tennessee}
  \country{USA}
  \postcode{37921}
}

\author{Weizi Li}
\email{weizili@utk.edu}
\orcid{0000-0002-0780-738X}
\affiliation{%
  \institution{University of Tennessee Knoxville}
  \streetaddress{P.O. Box 1212}
  \city{Knoxville}
  \state{Tennessee}
  \country{USA}
  \postcode{37921}
}

\author{Sai Swaminathan}
\email{sai@utk.edu}
\orcid{0009-0009-5055-2632}
\affiliation{%
  \institution{University of Tennessee Knoxville}
  \streetaddress{P.O. Box 1212}
  \city{Knoxville}
  \state{Tennessee}
  \country{USA}
  \postcode{37921}
}


\renewcommand{\shortauthors}{Zahid et al.}

\begin{abstract}
 
Maintaining an active lifestyle is vital for quality of life, yet challenging for wheelchair users. For instance, powered wheelchairs face increasing risks of obesity and deconditioning due to inactivity,  Conversely, manual wheelchair users, who propel the wheelchair by pushing the wheelchair's handrims, often face upper extremity injuries from repetitive motions, with shoulder pain affecting 42-66\%. These challenges underscore the need for a mobility system that promotes activity while minimizing injury risk. Maintaining optimal exertion during wheelchair use enhances health benefits and engagement, yet the variations in individual physiological responses complicate exertion optimization. 
To address this, we introduce \model{}, a novel wheelchair system that provides personalized assistance based on each user’s physiological responses, helping them maintain their physical exertion goals. Unlike conventional assistive systems focused on obstacle avoidance and navigation, \model{} integrates real-time physiological data—such as heart rate and ECG—with wheelchair speed to deliver adaptive assistance. Using a human-in-the-loop reinforcement learning approach with Deep Q-Network algorithm (DQN), the system adjusts push assistance to keep users within a moderate activity range without under- or over-exertion. We conducted preliminary tests with 10 users on various terrains, including carpet and slate, to assess \model{}’s effectiveness. Our findings show that, for individual users, PulseRide maintains heart rates within the moderate activity zone as much as 71.7\% longer than manual wheelchairs. Among all users, we observed an average reduction in muscle contractions of 41.86\%, delaying fatigue onset and enhancing overall comfort and engagement. These results indicate that PulseRide offers a healthier, adaptive mobility solution, bridging the gap between passive and physically taxing mobility options.

\end{abstract}


%
\received{6 January 2025}
\received[revised]{31 March 2025}
\received[accepted]{30 June 20259}
\maketitle
\section{Introduction}
\begin{figure*}[t!]
  \includegraphics[width=0.99\textwidth]{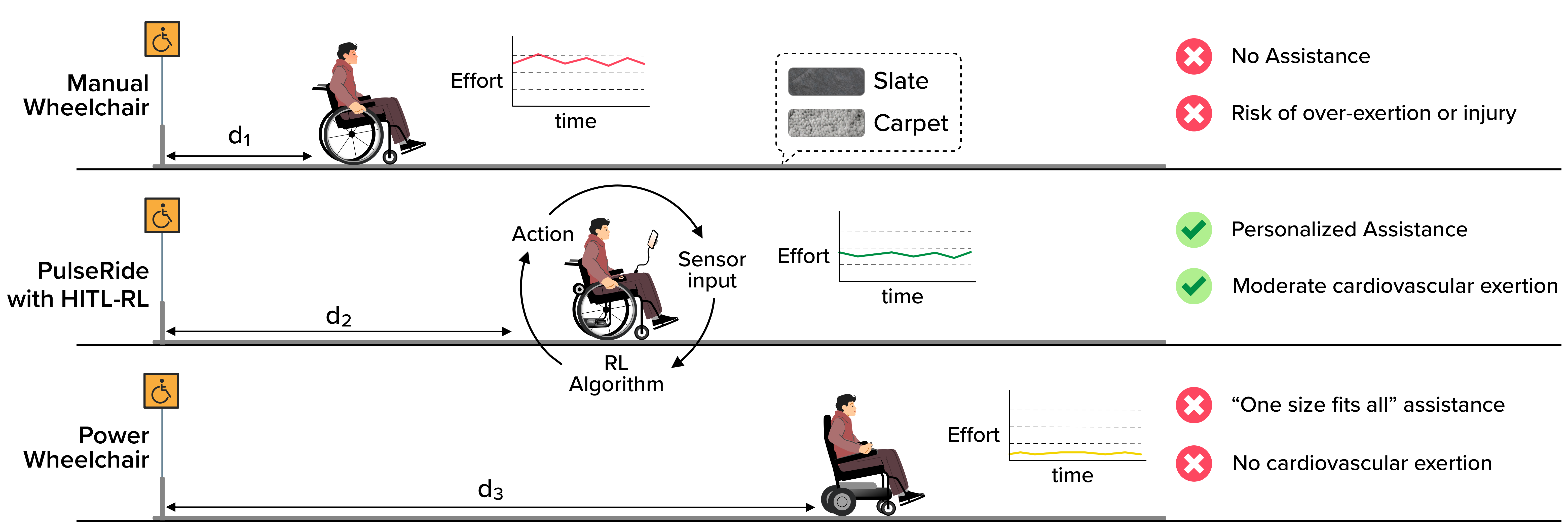}
  \caption{Comparison of wheelchair assistance paradigms. Manual wheelchair (top) provides no assistance and requires maximum user effort, leading to early onset of fatigue and risk of over-exertion. Power wheelchair (bottom) offers fixed assistance but discourages physical activity i.e., complete reliance on motorized propulsion. PulseRide (middle) provides personalized adaptive assistance, balancing user's physical effort (to maintain moderate cardiovascular exertion) and mobility needs. For the same effort budget, as \model{} delays the onset of fatigue, users are able to cover more distance (d2>d1).}
  \label{fig:teaser}
  \Description[]{}
\end{figure*}
Global demand for assistive technologies is substantial and growing. The World Health Organization (WHO) estimates that 2.5 billion people need one or more assistive products, a figure projected to reach 3.5 billion by 2050~\cite{world2022global}. A significant portion of this need is related to mobility impairment, requiring assistive devices like canes, walkers, and wheelchairs~\cite{sehgal2021mobility}. In particular, 80 million people rely on wheelchairs for daily mobility~\cite{who_assistive_technology}, essential for both routine tasks (e.g., moving around the home) and broader social participation. While standard manual wheelchairs provide essential mobility assistance, they rely on user effort for propulsion. Carpets, rough terrain, and even small inclines or slopes increase the effort and energy cost of mobility~\cite{Holloway2015}. A growing body of evidence suggests that the repetitive stress of manual propulsion has serious health consequences. The continuous repetitive motion often leads to upper extremity injuries, with shoulder pain reported in 42\% \cite{Dalyan1999Mar} to 66\% of users~\cite{Fullerton2003Dec} and common injuries to the rotator cuff muscles~\cite{Akbar2010Jan}. Additionally, bilateral carpal tunnel syndrome is frequently associated with handrim manipulation and engagement~\cite{Asheghan2016May}. These injuries could be mitigated by reducing push frequency to decrease the repetitive strain associated with long-term wheelchair use~\cite{Boninger2005May}.

Power wheelchairs offer an alternative by providing motorized assistance, alleviating the physical strain of manual propulsion. However, they often provide fixed levels of support focused solely on mobility, addressing the basic requirement of independent movement from point A to B, but generally overlook the user's need for physical activity and its cardiovascular benefits~\cite{edwards2010survey, andrabi2022physical}. Physical inactivity disproportionately affects wheelchair users, contributing to obesity, deconditioning, and further health decline~\cite{Cooper1999Apr}. Additionally, power wheelchair users may experience limitations in community participation \cite{Giesbrecht2009May}, as reduced physical activity can hinder engagement in social settings, further impacting their well-being. A third option has emerged to address the limitations of conventional manual and power mobility: power assist wheels. For instance, systems such as SmartDrive, Yomper \cite{acekare_yomper_enjo} and e-motion \cite{alber_emotion} have shown to allow users to stroke the wheelchair pushrims to activate small, lightweight motors that drive the wheels for a brief period of time (seconds). Users must continue to stroke the pushrim as they would if they were propelling a conventional manual chair. This method requires less strength and endurance than manual wheeling and is especially helpful on inclines, uneven terrain, and carpeted surfaces. However, most power-assist systems still operate with a “one-size-fits-all” approach to assistance, lacking automatic adaptation to varying environmental conditions (e.g., slopes, ramps, gravel) and overlooking the variability in individual user capabilities and contexts. Assistance levels are typically adjusted manually using a knob with preset velocity settings, making these systems less responsive to real-time user and environment needs.

These limitations highlight a critical gap in assistive technology that could dynamically personalize assistance to match each user’s unique ability ~\cite{wobbrock2011ability}, physiological state and environmental context. Additionally, wheelchair users often have limited opportunities for physical activity, with manual propulsion remaining one of the few accessible forms of exercise~\cite{coulter2011development}. Integrating physical activity into daily routines, beyond structured leisure time, is increasingly recommended to promote holistic health benefits~\cite{cowan2022lifestyle}. Therefore, determining the appropriate level and timing of automatic assistance to meet both mobility and health goals remains an open question. Additionally, understanding how an autonomous system could achieve physical activity goals as part of daily mobility for wheelchair users is still underexplored. In response, we propose a new assistance paradigm for wheelchair users that integrates real-time physiological feedback with autonomous control to deliver personalized, exertion-aware assistance. As shown in Figure~\ref{fig:teaser}, this approach provides a balanced solution between manual wheelchairs that require maximum user effort and power wheelchairs that discourage physical activity. Our research focuses on the following questions: \begin{itemize} \item \textbf{RQ1}: How can we design an autonomous wheelchair system that promotes physical activity while supporting users’ daily mobility?  \item \textbf{RQ2}: How can we personalize assistance levels by dynamically adapting to users' physiological states? \item \textbf{RQ3}: How can we ensure that this adaptive assistance is effective across diverse environments? \end{itemize}

To address these research questions, we designed a novel, autonomous wheelchair system called PulseRide. PulseRide utilizes Human-in-the-Loop Reinforcement Learning (HITL-RL), integrating real-time data on users' physiological state including heart rate and electrocardiogram (ECG) signals, and wheelchair movement (speed) to dynamically adjust the timing and duration of automatic push assistance. This approach enables PulseRide to continuously balance physical support based on user exertion, adapting in real time to match individual needs and environmental conditions. To the best of our knowledge, PulseRide is the first wheelchair system to offer personalized, adaptive assistance that balances users' physical effort with their mobility needs as a part of their daily mobility.
Through a human subject study with 10 participants (diverse in sex, age, and body mass index) across two indoor surfaces (slate and carpet), we demonstrate the ability of PulseRide to improve: (1) time spent in the moderate heart rate zone by up to 71.7\% over manual wheelchair use, (2) reduction of muscle contraction by 41.86\% and 24.94\% across multiple environments (slate and carpet), and (3) fatigue resistance, delaying onset of fatigue. These results demonstrate that PulseRide successfully achieves its core objectives of providing a personalized, exertion-aware daily mobility solution that promotes physical health.

\section{Background and Related Work}
\subsection{Physical Activity and Wheelchair Users}
Current research strongly supports the beneficial effects of physical activity (PA) on human health \cite{liao2018just} and well-being \cite{booth2000waging, haskell2007physical}. Physical activity has been shown to help prevent chronic conditions such as cardiovascular disease, diabetes, cancer, hypertension, obesity, depression, and osteoporosis \cite{warburton2006health, kell2001musculoskeletal, warburton2001effects}. For wheelchair users in particular, PA is especially valuable for maintaining or improving physical capacity \cite{muraki2000multivariate, janssen1996changes}, contributing to an increased quality of life \cite{heath19978,hwang2024acm}. Both the Centers for Disease Control and Prevention and the American College of Sports Medicine recommend a minimum of 30 minutes of moderate-intensity activity daily \cite{pate1995physical} to achieve health benefits such as decreased depression and anxiety \cite{office2000us, vuillemin2005leisure}. Another guideline suggests 150 minutes of moderate-intensity activity per week \cite{nash2012evidence}. These recommendations are particularly important for wheelchair users, as they help reduce the risk of secondary health problems that could further limit functional independence \cite{pate1995physical}. However, despite these clear benefits, physical inactivity is especially prevalent among wheelchair users \cite{office2000us, heath19978, tolerico2007assessing}. Many wheelchair users struggle to engage in the recommended levels of health-promoting exercise \cite{bussmann2009ambulatory, fougeyrollas2000long}. For wheelchair users, opportunities to be physically active are often limited, with manual wheelchair propulsion remaining one of the few accessible forms of PA and exercise \cite{coulter2011development}. To address this gap, researchers have suggested integrating PA into daily routines \cite{vuillemin2005leisure, rejeski1996physical}, moving beyond traditional structured leisure-time exercise.

This shift has led to a focus on "Lifestyle Physical Activity" (LPA) \cite{cowan2022lifestyle}, which emphasizes incorporating moderate-to-vigorous PA into everyday life rather than relying on structured exercise alone \cite{piercy2018physical}. LPA involves self-selected activities—including leisure, occupational, and household tasks—that meet at least moderate intensity levels and can be part of daily routines, whether planned or unplanned \cite{dunn1998lifestyle}. However, LPA has not yet been clearly defined for wheelchair users, and personalized PA schemes for this group remain underdeveloped. Moreover, measuring PA in wheelchair users has proven challenging due to variability among individuals and the complexity of assessing their PA levels. In addressing these measurement challenges, researchers have explored various approaches. Coulter et al. \cite{coulter2011development} used tri-axial accelerometers mounted on wheelchairs to measure physical activities through motion, a method that has become common in the literature \cite{nightingale2017measurement, gendle2012wheelchair}. However, these devices do not directly capture the user's actual activity levels through on-body sensors, which could provide more detailed and specific information about the user's activity.

In this work, we aim to bridge the gap between existing physical activity recommendations and their practical implementation for wheelchair users by developing a system that integrates real-time physiological feedback with adaptive mobility assistance. While prior studies have explored activity measurement using wheelchair-mounted accelerometers \cite{coulter2011development, nightingale2017measurement}, these approaches focus primarily on motion-based metrics and lack real-time, personalized adjustments based on user-specific physiological states. Our work addresses this limitation by incorporating data such as heart rate and ECG to provide personalized, exertion-aware assistance. By combining physiological sensing with Human-in-the-Loop Reinforcement Learning (HITL-RL), we enable dynamic adaptation of wheelchair push assistance to meet individual user needs. This personalized approach not only encourages sustained physical activity but also supports users in achieving moderate exertion levels.

\begin{figure*}[t!]
    \centering
    \includegraphics[width=0.85\linewidth]{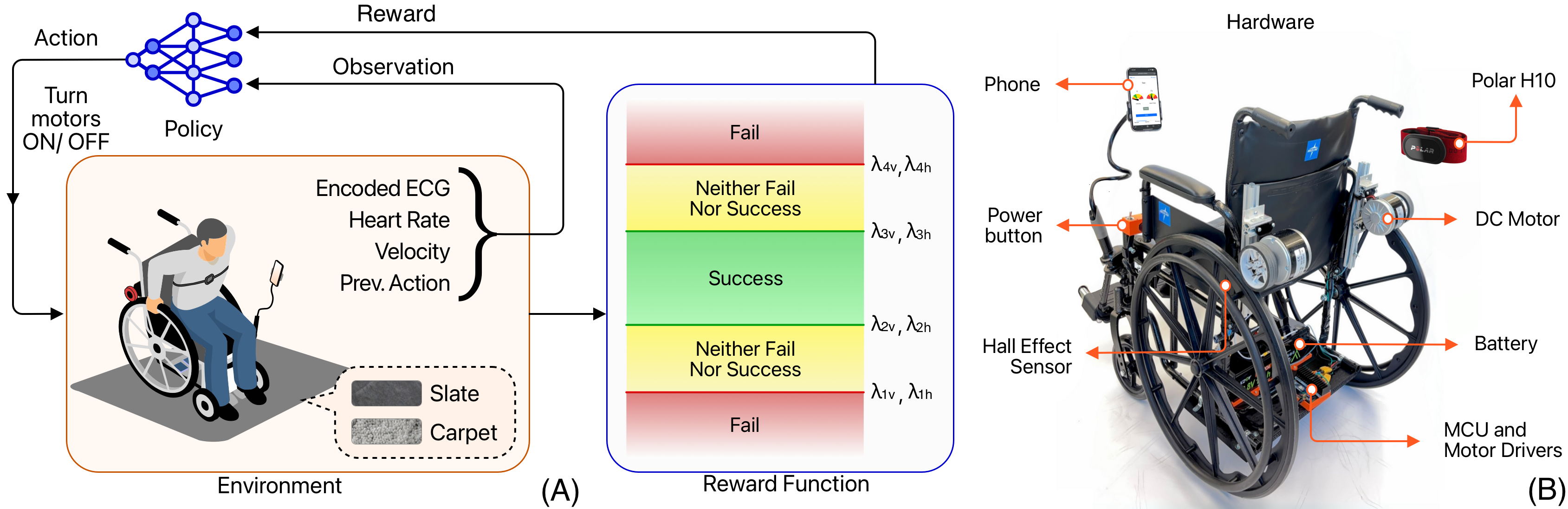}
    \vspace{-1mm}
    \caption{(A) PulseRide’s reinforcement learning framework, using inputs such as encoded ECG, heart rate, velocity, and previous action to adapt motor activation via a reward function that encourages moderate exertion. (B) PulseRide hardware setup, including the Polar H10 heart rate sensor, phone interface, DC motors, hall effect sensor, motor drivers, MCU, power button, and battery.}
    \label{fig:exp_setup}
    \Description[]{}
    \vspace{-4mm}
\end{figure*}

\subsection{Human-in-the-Loop Reinforcement Learning for Smart Wheelchair}
\label{related_HITL_RL}

Traditional assistance methods often rely on predetermined models \cite{feng2019data} or fixed thresholds \cite{wan_2014_ACC}, which struggle to capture the dynamic nature of user fatigue and recovery patterns \cite{liao_2020_acm}. While current AI approaches excel at analyzing health data, they primarily focus on monitoring and prediction rather than real-time adaptation of assistance. In this context, Reinforcement Learning (RL) offers a promising solution for real-time adaptation, though its application in smart wheelchairs has primarily focused on autonomous navigation tasks \cite{chatzidimitriadis2022deep, ryu2021development, lu2021assistive}. Goli et al. \cite{goli2013ieee} demonstrated the importance of human feedback in optimizing navigation assistance, yet the integration of physiological indicators within RL frameworks for physical activity support remains largely unexplored.

Human-in-the-Loop Reinforcement Learning (HITL-RL) extends traditional RL by incorporating explicit human input into the learning process. While conventional RL systems learn through environmental rewards alone, HITL-RL leverages direct human input to shape the policy, particularly valuable when the reward function cannot fully capture complex human preferences and comfort levels \cite{escobedo2013multimodal, tomari2012development}. This approach is especially relevant for wheelchair assistance, where user comfort, fatigue levels, and physical activity goals must be balanced dynamically. The integration of human feedback in HITL-RL offers two key advantages for wheelchair assistance systems. First, it enables the system to learn from both immediate physiological responses (like heart rate and ECG) and explicit user preferences, creating more personalized assistance policies. Second, human guidance helps address the exploration-exploitation challenges in RL \cite{luo2024ieee}, allowing safer and more efficient learning in real-world scenarios. This framework provides the foundation for adaptive assistance that can simultaneously support both mobility requirements and physical activity goals, while maintaining the personalization benefits of AI-driven approaches.

\section{System Design}
\label{system_design}

In this section, we provide an overview of \model{}'s design and a walkthrough of its operation. We then describe how Human-in-the-Loop Reinforcement Learning (HITL-RL) integrates real-time data on users' physiological state and wheelchair movement to dynamically adjust the level of automatic push assistance. Finally detailing other the technical aspects such neural network model architectures. 

\subsection{\model{}Hardware and Operation}

\model{} is designed using a standard off-the-shelf manual wheel-chair from Medline (Model Drive Medical SSP118FA-SF). We motorized the wheelchair with two 250W 24V DC brushed motors powered by two 12V lithium batteries, mounted on aluminum extrusions behind the wheelchair (Figure 2B). The motors are connected to the wheelchair tires via grip wheels, allowing for powered assistance when needed. We chose brushed DC motors over stepper motors due to their low resistance when turned off, which enables users to manually propel the wheelchair without additional resistance. For monitoring and sensing, \model{} employs an array of sensors. To measure human physiological responses, we use a Polar H10 sensor for ECG and heart rate. To track wheelchair velocity, we installed a hall effect sensor along with 18 magnets on the wheel rims for RPM calculation, with each increment representing a rotation of 20 degrees (360°/18). The system supports a maximum load of 113 kg and includes a speed limit control that caps the maximum speed at 2.5 m/s to ensure safety. To further enhance safety, we incorporated a power cut-off switch near the armrest and a remote kill switch, allowing for immediate shutdown if necessary. The total cost of the attachable hardware, including the manual wheelchair, is under \$500, making it an affordable solution.

Upon starting the \model{} system, the user connects to the wheelchair via their mobile phone, where they are presented with an interface showing important safety information. Next, before fully operating \model{}, the user completes a three-step setup process. First is a pretraining phase, where the user propels the wheelchair for 135 seconds at three perceived exertion levels (low, medium, and high), specific to them. During the pre-training phase, \model{} calculates user-specific heart rate and velocity thresholds, which are represented as activity zones to guide the user. These zones are visualized with green representing the moderate activity zone, yellow representing transition zones (between moderate and extreme), and red representing both low and high extreme activity zones (Figure 2A). The user then pushes the wheelchair for 20 minutes across these zones as instructed by our AI algorithm, receiving prompts on the mobile interface, such as “go slow” or “go fast,” as directed. After completing this AI-guided training, the user enters the testing mode and can operate \model{} with adaptive assistance. 

\begin{figure*}[t!]
    \centering
    \includegraphics[width=0.95\linewidth]{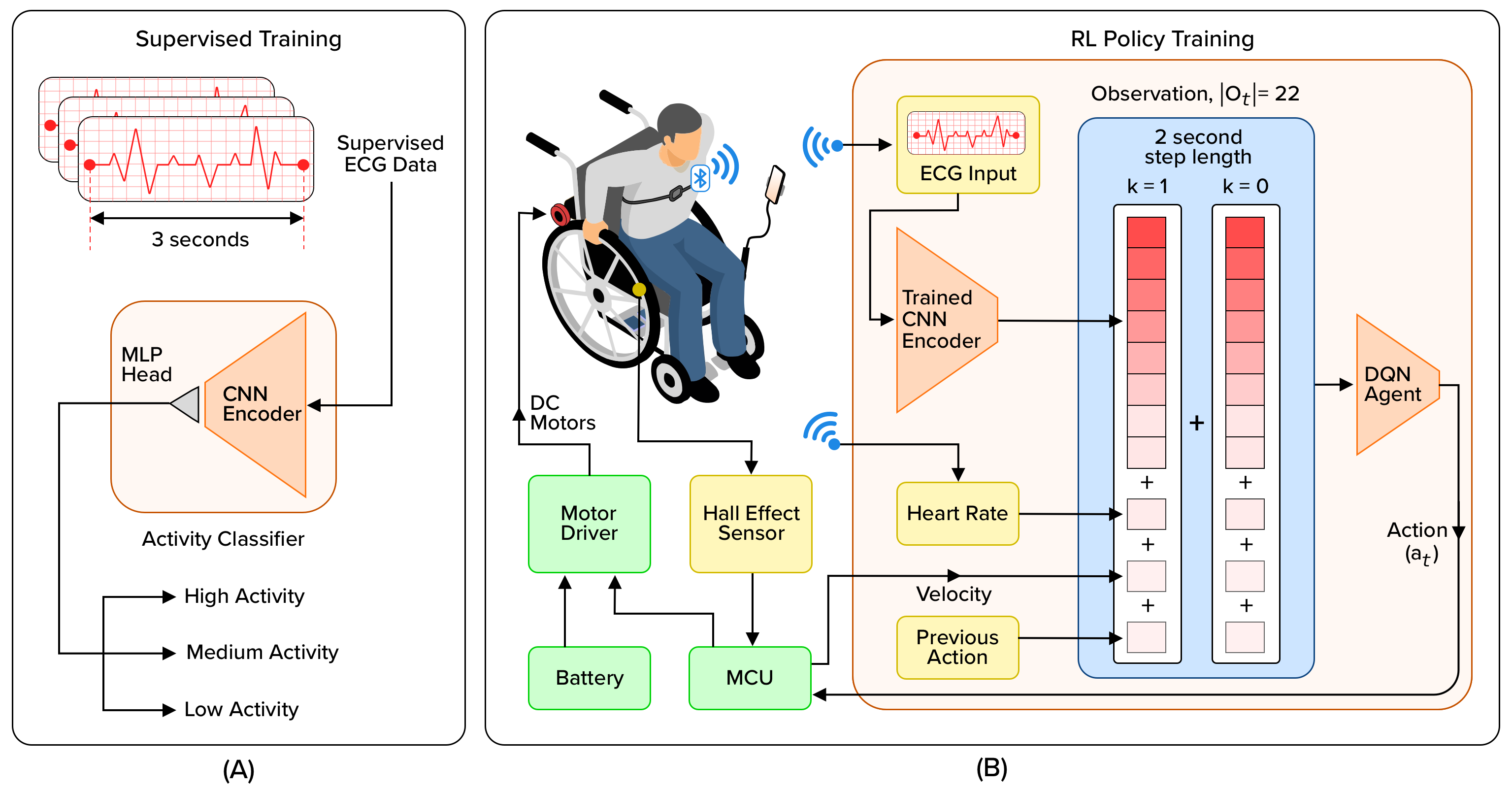}
    \vspace{-1mm}
    \caption{\textbf{(A)} Activity classifier consisting of a Convolutional Neural Network (CNN) encoder and a Multi-Layered Perceptron (MLP) head, trained on a supervised dataset to classify ECG signals into three activity classes (High, Medium, and Low). \textbf{(B)} The realtime, high-fidelity ECG signals are input to the trained CNN Encoder (frozen during RL policy training) which feeds the encoded output into the observations of the RL pipeline. Additionally, the heart rate from the Polar heart rate monitor, the velocity from the hall effect sensor and the previous actions are also added as observations.  The DQN Agent passes its actions to the wheelchair microcontroller unit (MCU) corresponding to either turn ON or turns OFF the motor.}
    
    \label{fig:trained_cnn}
    \Description[CNN Encoder]{Activity classifier consisting of a Convolutional Neural Network (CNN) encoder and a Multi-Layered Perceptron (MLP) head, trained on a supervised dataset to classify ECG signals into three activity classes (High, Medium, and Low). (B): The trained CNN Encoder (frozen) is incorporated in the observations of the RL pipeline to encode high-fidelity ECG signals. Additionally, the heart rate, velocity and previous actions are input to the DQN Agent.}
    \vspace{-4mm}
\end{figure*}

During the training phase, \model{} uses representation learning through neural networks in two distinct stages: ECG signal encoding and the RL policy as shown in Figure \ref{fig:trained_cnn}. The ECG encoder, trained to classify activity levels based on ECG input (shown in Figure \ref{fig:trained_cnn} (A)), learns to extract meaningful physiological patterns from the signals. HITL-RL then processes these encoded patterns along with heart rate and velocity data to determine when and how long to provide assistance to maintain moderate exertion levels. This approach allows \model{} to automatically identify relevant patterns in user behavior and physiological responses that impact their ability to propel the wheelchair. 
Notably, there is no explicit programming of which features \model{} should analyze; potential factors include fatigue patterns, muscular strength, and heart rate response.

\subsection{Reinforcement Learning Approach}
\label{hitl_rl}
We now go over the artificial intelligence (AI) algorithm used by PulseRide for providing dynamic assistance. As introduced in Section \ref{related_HITL_RL}, we are using a variation of Reinforcement Learning called  Human-in-the-Loop Reinforcement Learning (HITL-RL). 

We formalize our approach using a Markov Decision Process, represented as the tuple $(\mathcal{S}, \mathcal{A}, \mathcal{P}, \mathcal{R}, \gamma)$. Here, $\mathcal{S}$ denotes the set of states, capturing all possible configurations of the environment and user; $\mathcal{A}$ is the set of actions the agent can take; $\mathcal{P}(s', r | s, a)$ defines the environment dynamics, representing the probability of transitioning to state $s'$ and receiving reward $r$ after taking action $a$ in state $s$; $\mathcal{R}(s,a)$ is the reward function, which assigns a numerical value to each state-action pair based on the desirability of that action in that state; and $\gamma$ is the discount factor that prioritizes immediate rewards over distant future rewards. The agent with a policy $\pi_{\omega}$, parameterized by $\omega$, aims to maximize the cumulative discounted rewards. The agent's actions are determined based on the Q-value function, which is iteratively updated according to the equation:
\begin{equation} 
    Q_{(s, a)} = \left(1-\alpha\right)Q_{\left(s, a\right)} + \alpha \left[ R_{\left(s,a\right)} + \gamma \max_{a'} Q_{(s', a')} \right], 
\end{equation}

\noindent where $\alpha$ represents the learning rate and $a'$ denotes the successor action of $a$. In scenarios with large state spaces, function approximation techniques, such as deep neural networks are preferred due to their ability to generalize and ensure convergence. We implement Deep Q Networks (DQN)~\cite{mnih2015human} as our RL algorithm, incorporating two crucial mechanisms: experience replay and target network. Experience replay maintains a memory buffer of observed state-action-reward transitions (size=120), from which we uniformly sample batches during training. This approach reduces variance in the learning process and addresses the non-stationarity of data distribution during policy updates. The target network serves as a separate Q-value estimator whose parameters are updated every timestep by 5\% ($\tau$) from the learned policy network, stabilizing the training process and preventing divergence of the policy learned by the agent. 


During training, the policy learns the timing (when to assist) and duration (for how long) while relying on the user for stopping and directional control. As designed, the \model{} policy cannot apply brakes or reverse movement. Because human responses are introduced into the control loop, an uncertain learning environment is created due to the inherent variability of such responses (e.g., when the policy reduces assistance by applying action 0 over multiple time steps, expecting increased user effort, the user might not respond as anticipated). This uncertainty introduces additional complexity to the exploration process, which is crucial for successful policy learning, as during training the policy needs to be sufficiently exposed to diverse state-action pairs to learn optimal behaviors~\cite{sutton2018reinforcement}. To systematically address these challenges, we implement a two-part exploration strategy during training that combines algorithmic exploration with structured human participation. At the algorithmic level, we employ an $\epsilon$-greedy approach with exponential decay which produces noisy actions from the DQN algorithm. At the human level, we implement a structured training protocol where users deliberately introduce periods of varying activity levels to expose the policy to different physiological states. Importantly, users are not required to follow any such protocol after obtaining a trained policy. To further stabilize the training process, we employ smooth L1 loss and gradient clipping to prevent large policy updates.

\subsection{Electrocardiogram Encoder}
We encode the Electrocardiogram (ECG) signals using a trained Convolutional Neural Network (CNN) before passing them as input to the RL agent (Figure~\ref{fig:trained_cnn} (B)). To leverage transfer learning, we first train the CNN on an adjunct supervised learning task of classifying ECG intervals into Low, Medium, and High activity levels (Figure~\ref{fig:trained_cnn} A). This training allows the encoder to learn general ECG features that are relevant for the RL task, even though the RL agent doesn't explicitly perform activity classification. After training, we freeze the CNN's weights and use it to transform three-second ECG intervals of length 390 (sampled at 130Hz) into concise vectors of eight dimensions, as shown in Figure~\ref{fig:trained_cnn} (B). 
By encoding the ECG signal into a vector of size eight, we strike a balance between providing the RL agent with the rich, high-fidelity ECG signal of size 390 and the simple, low-fidelity heart rate measurement of size one. This approach provides more nuanced physiological information than heart rate alone, while avoiding the computational complexity of processing entire raw ECG signals. Hence, to train the encoder, we collected data from four participants separate from our main participant pool and split the dataset into 80\% training and 20\% validation sets. After 500 epochs of training, the CNN encoder achieves 100\% training accuracy and 87.5\% validation accuracy. We then remove the final MLP head from the model, allowing it to encode ECG signals into eight-dimensional vectors that capture meaningful physiological features. Table~\ref{tab:hyperparameters} provides a comprehensive list of hyperparameters used in the CNN encoder training.

\begin{table}[htb]
\centering
\footnotesize
\begin{tabular}{llc}
\toprule
\textbf{Model} & \textbf{Hyper-parameter} & \textbf{Value} \\
\midrule
 & Batch size & $16$ \\
 & Learning rate & $0.0001$ \\
 & Number of epochs & $500$ \\
 CNN & Optimizer & Adam \\
 Encoder& CNN Kernel size & $5$ \\
 & Number of training samples & $303$ \\
 & Loss function & Cross-Entropy \\
\midrule
 & Replay buffer size & $120$ \\
 & Batch size & $32$ \\
 & Discount factor ($\gamma$) & $0.99$ \\
 & Optimizer & Adam \\
 & Exploration rate (initial $\epsilon$) & $0.99$ \\
 HITL-RL & Exploration rate (final $\epsilon$) & $0.05$ \\
 & Activation function & ReLU \\
 & Optimizer & Adam~\cite{kingma2014adam} \\
 & Gradient clipping & $[-100, 100]$ \\
 & Target Network update ratio ($\tau$) & $0.005$ \\
 & Number of training episodes & $30$ \\
\bottomrule
\end{tabular}
\caption{Training parameters and hyper-parameters for CNN Encoder and HITL-RL.}
\label{tab:hyperparameters}
\vspace{-6mm}
\end{table}

\subsection{Markov Decision Process}
The main components of our Markov decision process are outlined below:

\textbf{Observation:} To sufficiently capture the dynamic behavior of the system, we represent observations with heart rate, velocity, previous action, and the output of the trained CNN encoder. Further, to capture transitions across multiple steps, we stack information from the two previous seconds to compose an observation in each time-step. This 2-second time-step (Figure~\ref{fig:trained_cnn} (B)) enables the agent to consider recent historical data when making decisions.

\textbf{Action:} The actions are binary i.e., either turn the two motors attached to the wheels ON (pushing \model{} forward) or OFF.

\textbf{Reward function:} To facilitate rapid learning, we employ a hybrid approach that incentivizes high-level desired outcomes (such as maintaining heart rate below the user's perceived over-exertion range) while simultaneously discouraging undesired actions (such as turning the motor ON when the system is already at high velocity). The complete reward function, shown in Algorithm~\ref{alg:reward}, incorporates user-specific empirical thresholds ($\lambda$ values) obtained during the pre-training phase, where users follow a specific protocol (defined in Section ~\ref{procedure}) of pushing \model{} at their own perceived low, medium, and high intensities. This approach ensures personalized adaptation to each individual's capabilities and limitations. Specifically, the function utilizes thresholds for both velocity ($\lambda_{1v}$, $\lambda_{2v}$, $\lambda_{3v}$) and heart rate ($\lambda_{1h}$, $\lambda_{2h}$, $\lambda_{3h}$) to define boundaries, as shown in Figure~\ref{fig:exp_setup}, for acceptable (`success' zone shown in green and `neither success nor failure' zone shown in yellow) and unacceptable (`failure' zone shown in red) ranges. Similar reward functions are commonly used in DC motor control problems~\cite{poudel2022learning}.


\begin{algorithm}[htb]
    \caption{Reward Function}
    \begin{algorithmic}
    
        \State{\texttt{\# Calculate distances}}
        \State{$vel\_dist \leftarrow \max(0, \text{velocity} - \lambda_{3v})$}
        \State{\textbf{if} $\text{velocity} < 0$ \textbf{then}}
        \State{\quad\quad$vel\_dist~\leftarrow ~max(0, |\text{velocity}| - |\lambda_{2v}|)$}
        \State{\textbf{end if}}
        \vspace{1mm}
        \State{$heart\_dist \leftarrow \max(0, \text{heart rate} - \lambda_{3h})$}
        \State{\textbf{if} $\text{heart rate} < 0$ \textbf{then}}
        \State{\quad\quad$heart\_dist \leftarrow \max(0, |\text{heart rate}| - |\lambda_{2h}|)$}
        \State{\textbf{end if}}
        \vspace{1mm}
        \State{}
        \State{\texttt{\# Calculate Reward}}
        \State{$reward \leftarrow -1 + 0.5 \cdot e^{-3 \cdot heart\_dist} + 0.5 \cdot e^{-2 \cdot vel\_dist}$}
        \vspace{1mm}
        \State{\textbf{if} ($\text{velocity} > \lambda_{1v}$) or ($\text{velocity} = vel\_cv$) or ($\text{heart rate} < \lambda_{2h}$) or ($\text{heart rate} > \lambda_{1h}$) \textbf{then}}
        \State{\quad\quad$reward \leftarrow -1.2$ \quad\texttt{\#~(failed = True) Terminate the episode}}
        \vspace{1mm}
        \State{\textbf{else if} ($\lambda_{2h} < \text{heart rate} \leq \lambda_{3h}$) and ($\text{velocity} > \lambda_{2v}$) \textbf{then}}
        \State{\quad\quad\textbf{if} $action = 1$ and $\text{velocity} > \lambda_{3v}$ \textbf{then}}
        \State{\quad\quad\quad\quad$reward \leftarrow -1 + e^{-(1.8 \cdot vel\_dist + 2.5 \cdot heart\_dist)}$}
        \State{\quad\quad\textbf{else}}
        \State{\quad\quad\quad\quad$reward \leftarrow 0.6 \cdot e^{-2 \cdot heart\_dist} + 0.6 \cdot e^{-2 \cdot vel\_dist}$}
        \State{\quad\quad\textbf{end if}}
        \vspace{1mm}
        \State{\textbf{else if} ($\lambda_{3h} < \text{heart rate} < \lambda_{1h}$) and ($\text{velocity} < \lambda_{3v}$) \textbf{then}}
        \State{\quad\quad$reward \leftarrow -1.4 + 0.5 \cdot e^{-3 \cdot heart\_dist} + 0.5 \cdot e^{-2 \cdot vel\_dist}$}
        \State{\textbf{end if}}
\end{algorithmic}
\label{alg:reward}
\end{algorithm}

\subsection{Model Architectures} The Deep Q Network (DQN) policy network and encoder are both Convolutional Neural Networks (CNNs), which are effective in processing physiological inputs~\cite{polomarco2023ecg}. The encoder architecture contains three ConvNormPool blocks that progressively process the single-channel input through 256, 128, and 64 channels. Within each ConvNormPool block, a residual connection adds the output of the first convolutional layer to the output of the third convolutional layer before normalization and Swish activation. This skip connection enhances feature preservation and facilitates better gradient flow during training~\cite{he2016deep}. After each block, a max pooling layer with kernel size 2 halves the dimensions of the feature maps. The output of three ConvNormPool blocks is passed through an adaptive average pooling layer and flattened before being passed through a fully connected layer that maps the 64 input features to eight features followed by a final linear layer (MLP head) maps these eight features to three output classes. Whereas in the HITL-RL policy network, the CNN architecture is designed to process physiological data by expressing the locality in time as locality in space (by stacking two consecutive timestep inputs), enabling the network to detect patterns in the temporal transitions of the data. The network takes consecutive velocities, voltages, encoded ECG signals, and heart rates as input. The policy network consists of two convolutional layers that expand the single input channel first to 8 and then to 16 channels, each followed by ReLU activation. The convolutional feature maps are then flattened and passed through three fully connected layers that progressively reduce the dimensionality from 832→128→32 neurons with ReLU activation after each layer. The final output layer produces two values corresponding to the Q-values for each action. The encoder and policy network have approximately one million and 110,000 total trainable parameters respectively.

\section{Methodology}
We now go over our experimental validation of the \model{} system, designed to evaluate its effectiveness in providing personalized, adaptive push assistance.
\subsection{Participants}
\label{participants}
We recruited 10 participants without disabilities, aged 19-30 (M = 22.8, SD = 3.65), seven participants identified as male and three as female. Participants had an average height of 67.3 inches (SD 4.92) and an average BMI of 23.63 (SD 4.27). Recruitment was conducted through university channels and social media. The inclusion criteria required participants to be physically able of propelling a manual wheelchair and providing informed consent. During recruitment, participants completed a questionnaire about their ability to engage in moderate physical activity and any relevant medical advice they had received regarding exercise. We excluded individuals with pre-existing cardiovascular or musculoskeletal conditions, as well as those with cognitive impairments that could prevent them from understanding the study tasks. Selected participants were compensated at a rate of \$20 per hour and the experiments took approximately 1.5 hours. This research protocol was reviewed and approved by the university ethics and review board.

The decision to conduct this study with the participants without disabilities rather than wheelchair users was made after consideration of two factors. Firstly, as this is an initial investigation of \model{}, we prioritized safety and sought to minimize potential risks to participants. Individuals without disabilities could more easily adapt to and recover from the physical demands of the study, allowing us to gather preliminary data on the system's effectiveness without exposing vulnerable populations to untested technology. Secondly, using participants without prior wheelchair experience helped us test how well our system adapts to new users. Since all participants started with the same level of wheelchair skills (none), any differences in their performance would likely be due to our system's effectiveness, not their prior abilities. While we acknowledge the limitations of this approach in terms of immediate real-world applicability, it serves as a crucial first step in its development. Further discussion of this consideration is presented in Section~\ref{able_bodied}.
\subsection{Procedure}
\label{procedure}
In the experiment procedure, participants start by receiving a brief introduction to the system, followed by a hands-on demonstration and a trial run. Then, they rest for a short rest period to establish their baseline heart rate. Next, the human-in-the-loop reinforcement learning (HITL-RL) policy for each participant is trained in two consecutive stages: 1) a pre-training phase, which lasts for two minutes and 15 seconds, and 2) a training phase which takes 20 minutes. The training is followed by test period of five minutes in two different environments.
\subsubsection{Pre-Training:} During the pre-training stage, the RL policy is turned off, and participants complete three consecutive 45-second intervals, pushing the wheelchair at their perceived moderate and transition (between moderate and extreme) activity levels. These intervals serve multiple purposes that enable system personalization. First, the heart rate range (min, max) is determined. Second, heart rate and velocity statistics (mean and standard deviation) used for normalizations are determined. And third, user-specific heart rate and velocity thresholds ($\lambda$ values used in the reward function of the training phase) that define the boundaries between moderate, transition, and extreme intensity zones are established. In Figure~\ref{fig:exp_setup}, these zones are visualized with green representing moderate activity zone shown as `success', yellow representing transition zones (between moderate and extreme) shown as `neither success nor failure', and red representing both low and high extreme activity zones shown as `failure'. 
\begin{figure}[htb]
    \centering
    \includegraphics[width=0.99\linewidth]{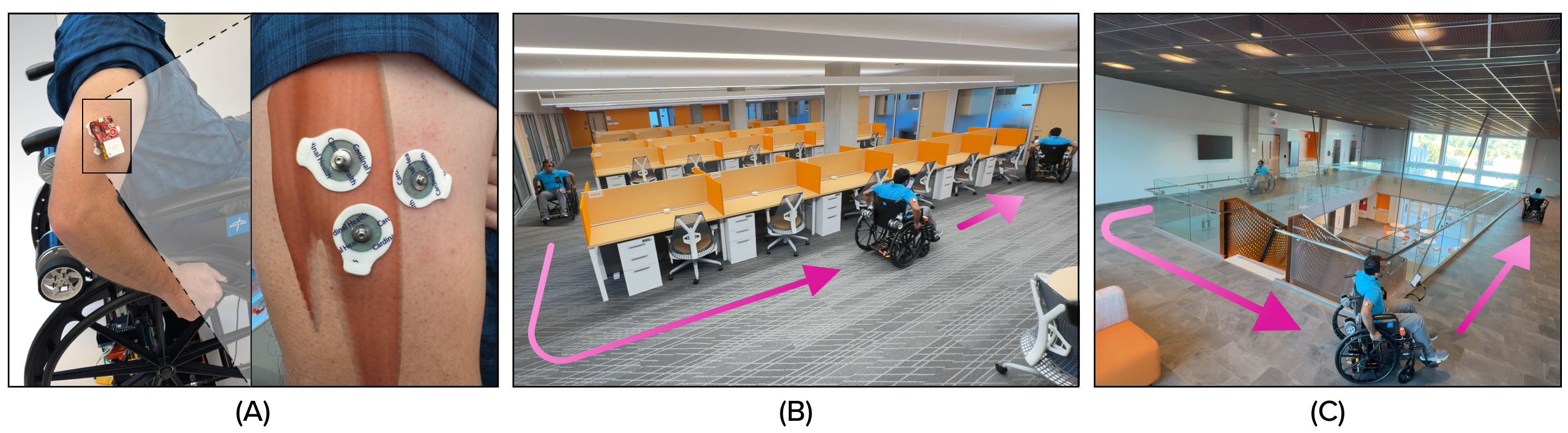}
    \caption{\textbf{(A)} Placement of electromyography (EMG) sensor electrodes on the lateral head - triceps brachii muscle of the right arm. \textbf{(B)} Indoor testing path on a carpeted floor. Users followed a closed-loop path in both environments with different coefficients of friction. The carpet surface exhibits a higher coefficient of friction than slate, requiring greater propulsion effort from users during wheelchair operation. \textbf{(C)} Indoor testing path on a slate floor.}
    \label{fig:environments}
    \Description[]{}
\end{figure}
\subsubsection{Training:} During training, participants are tasked with maintaining \model{} velocity within their personalized moderate velocity zone, established during pre-training. The moderate velocity zone is visually represented by the green zone in the center of the user interface gauges. Two real-time analog needles continuously indicate instantaneous heart rate and velocity. As described in Section~\ref{hitl_rl}, to evaluate system response outside the moderate zones, participants receive prompts in the user interface to alter their speed and activity intensity. 
 The prompt at training episode 15 is 'Go slower' and the prompt at training episode 25 is 'Go faster'. The complete training period comprises of 30 episodes and each episode contains 20 timesteps of 2 seconds duration. Notably, all training sessions were conducted only on a slate surface, with participants following a closed-loop path of approximately 50 meters perimeter (Figure~\ref{fig:environments}).
\subsubsection{Testing: } Following training, each participant underwent a test phase consisting of five episodes (30 timesteps each). Similar to training, participants followed the same closed loop paths but on two different surfaces - slate and carpet flooring (Figures~\ref{fig:environments}B \& ~\ref{fig:environments}C). The carpeted surface, with its higher coefficient of friction, required more physical effort for wheelchair propulsion compared to the smooth slate surface. Participants were tasked with maintaining their personalized moderate velocity `green' zone established during pre-training. To collect data for validation, EMG sensors were placed on the triceps of participants' both arms using MyoWare 2.0 muscle sensor development kits, which include an Electromyography (EMG) muscle sensor, a wireless communication shield, and solid gel electrodes. We attached the electrodes along the longitudinal midline of each triceps head, parallel to the muscle fibers, avoiding placement directly on motor points and at the edges of the muscle to prevent crosstalk from adjacent muscles~\cite{Ohashi2022Sensors} as shown in Figure~\ref{fig:environments}A. The EMG sensor input measurement rate is 20Hz, further, to ensure sufficient data capture, we started the collection process earlier and stopped later than the actual test time. During analysis, the collected data was synchronized with the correct test timing. It is important to note that EMG data was only used for additional validation and analysis during test period and is not a part of the input to the RL agent, hence it is excluded from Section~\ref{system_design}.
\subsection{Measures and Analysis}Our experiments investigate \model{}'s effectiveness across multiple dimensions with comparisons to manual wheelchair use. Specifically, we study four physiological and task outcomes:
\subsubsection{Number of muscle contractions:} We quantify the muscular effort by counting the total number of muscle contractions. A muscle contraction is detected on the raw EMG signal using root mean square (RMS) as a threshold i.e., we detect a contraction event if the raw EMG exceeds the RMS value. Since both push and recovery (pulling back and cocking the arms for another push) movements trigger muscle contractions, they directly correlate with the number of times a wheelchair is pushed.
\subsubsection{User-Specific Personalization:} We evaluate \model{}'s adaptation to individual users by analyzing their heart rate and velocity distributions across different activity zones (moderate, transition, and extreme). Using Gaussian kernel density estimation, we construct continuous probability density functions of these physiological responses from observed data. The proportion of time spent in each zone, particularly the moderate activity zone, directly quantifies \model{}'s effectiveness in maintaining personalized activity levels.
\subsubsection{Generalization Across Environments:} We assess \model{} environmental adaptability by comparing physiological responses across two distinct surfaces: Slate (smooth) and Carpet (rough). The key measure is the consistency of heart rate zones across environments i.e., whether users maintain similar physiological states regardless of surface conditions. This comparison reveals the system's ability to adapt assistance levels to varying environmental demands while maintaining user comfort.
\subsubsection{Fatigue Reduction:} We evaluate \model{}'s ability to reduce fatigue by analyzing electromyography (EMG) signals compared to manual wheelchair use. As muscles fatigue, there is a shift in the EMG power spectrum toward lower frequencies, which manifests as a decrease in mean frequency (MNF) \cite{rodrigues2022emg}. To quantitatively assess fatigue delay, we conducted MNF analysis on raw EMG data by dividing the total testing time of 300 seconds into 30 equal 10 seconds segments. Then for each segment, we determined the mean frequency and calculated the best-fitted approximation of the MNF data across all timeslots, allowing us to track muscle fatigue progression throughout the testing period.
\section{Results}

\subsection{Personalization across users}

\begin{figure}
    \centering
    \includegraphics[width=\linewidth]{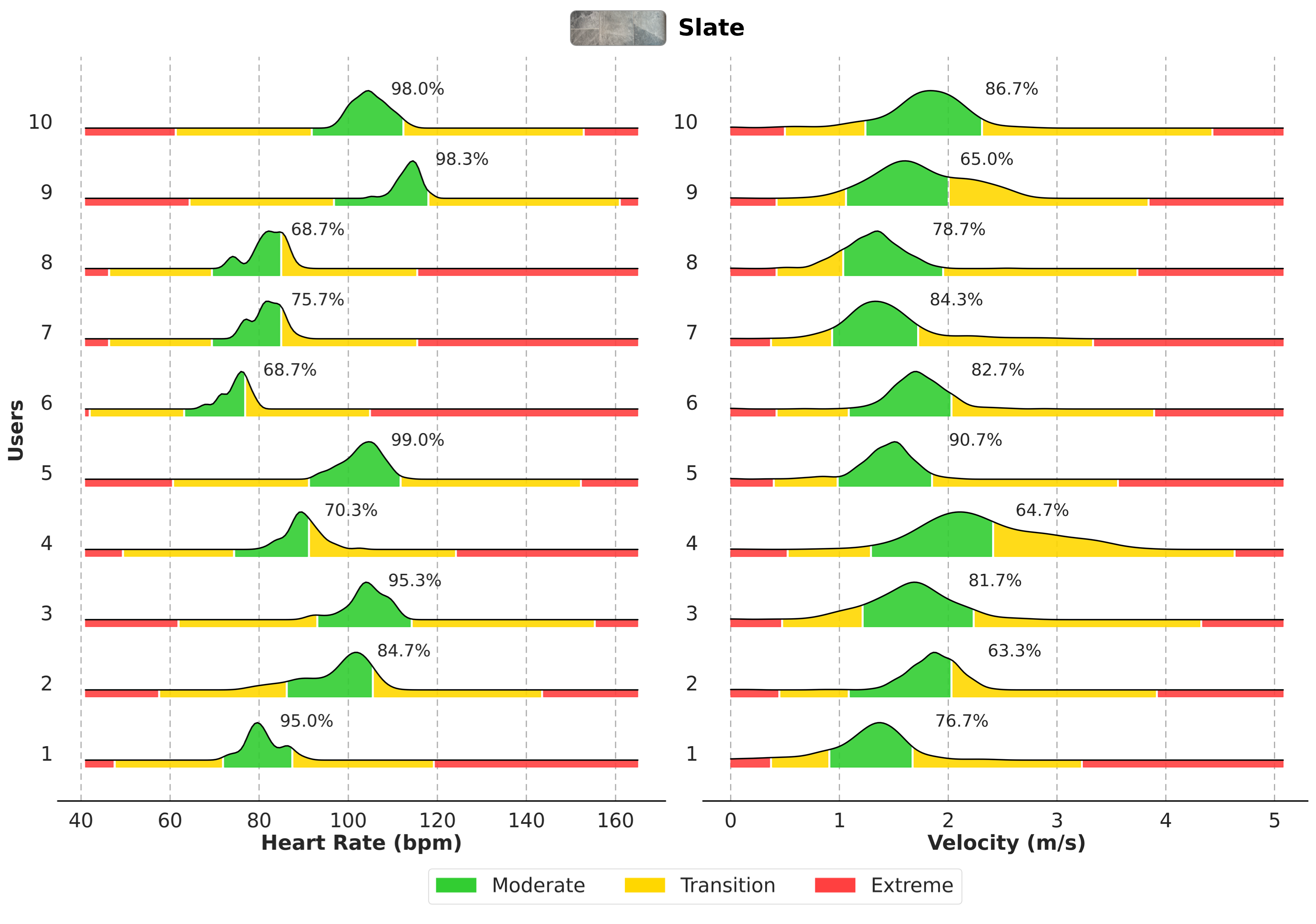}
    \caption{Heart rate and velocity distributions across users on Slate environment, estimated using Gaussian kernel on observed data. The x-axis presents each user's personalized ranges corresponding to moderate, transition and extreme activity levels. 
    }
    \label{fig:personalize_slate}
    \Description[]{}
    \vspace{-4mm}
\end{figure}

\begin{figure}
    \centering
    \includegraphics[width=\linewidth]{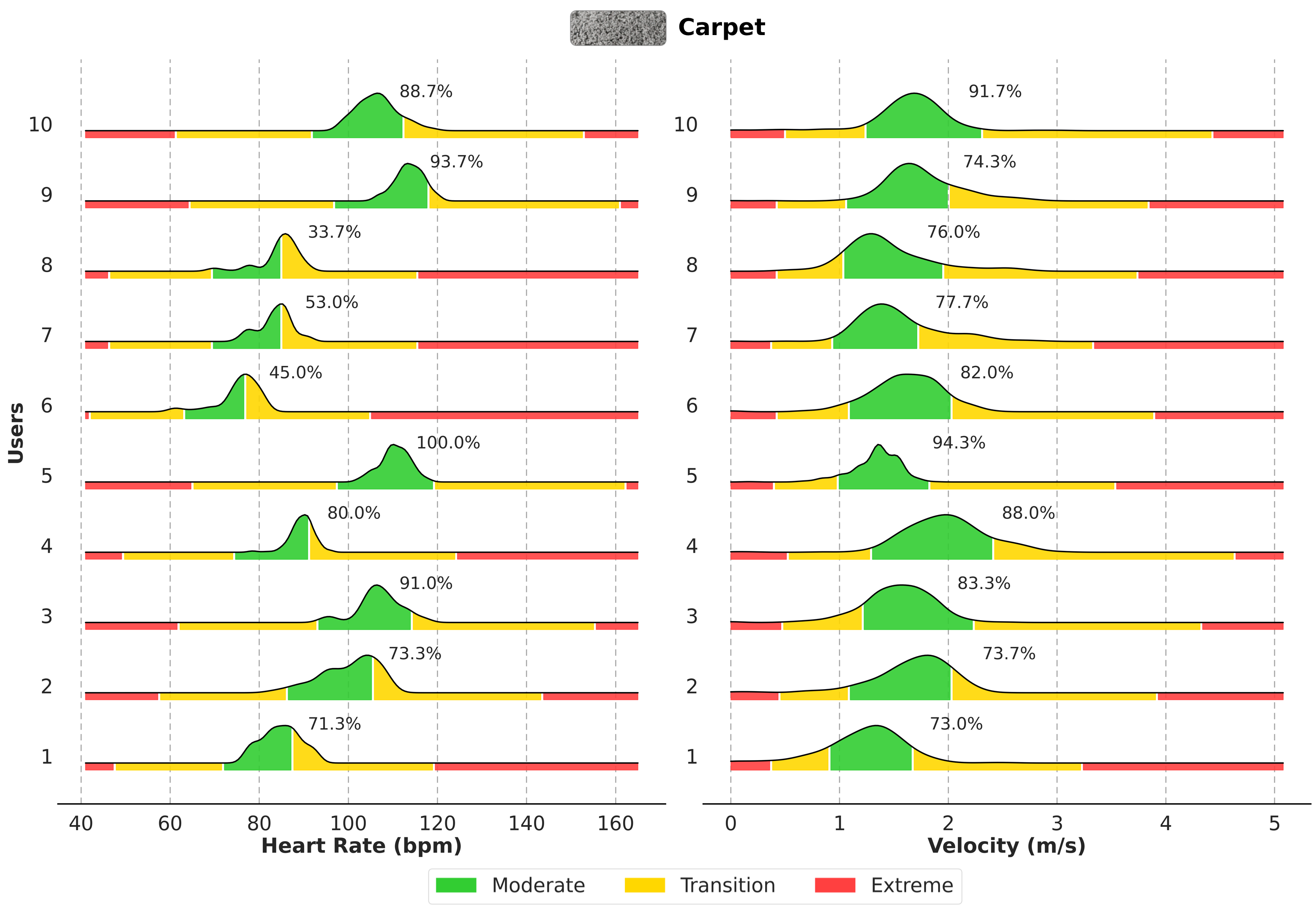}
    \vspace{-4mm}
    \caption{Heart rate and velocity distributions across users on Carpet environment, estimated using Gaussian kernel on observed data. The x-axis presents each user's personalized ranges corresponding to moderate, transition and extreme activity levels. 
    }
    \label{fig:personalize_carpet}
    \Description[]{}
    \vspace{-2mm}
\end{figure}

Our results indicate that \model{} effectively personalizes heart rate activity, helping users maintain heart rates within their individual moderate activity zones across all participants. As shown in Figure ~\ref{fig:personalize_slate} and Figure ~\ref{fig:personalize_carpet}, despite each user having a unique moderate heart rate zone, \model{} consistently enabled users to remain within these personalized zones. However, individual analysis reveals some outliers. For instance, Users 8, 7, and 6 spent only 33.7\%, 53\%, and 45\% of the time in the moderate heart rate zone, respectively, with a higher percentage of time in the transition zone toward the upper extreme heart rate zone. This deviation can be explained by examining the velocity distribution for these users, where they spent more amount of time in the transition zone toward higher velocities. Since \model{} does not have explicit braking capabilities, users may have been pushing themselves more in these instances.

\begin{figure}[h]
    \centering
    \includegraphics[width=1\linewidth]{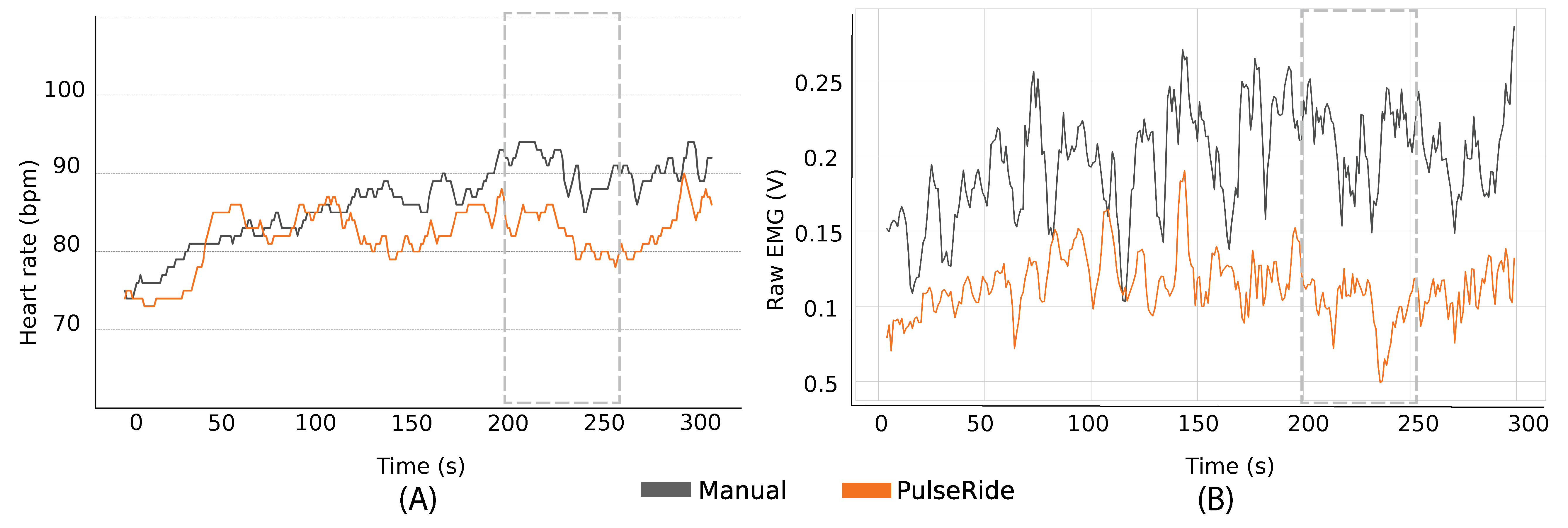}
    \caption{\textbf{A.} Heart rate trajectory for one user throughout the test experiment on Slate. \textbf{B.} EMG trajectory for one user throughout the test experiment on Slate. }
    \label{fig:hr_emg_traj}
    \Description[]{}
    \vspace{-2mm}
\end{figure}

Additionally, Figures \ref{fig:hr_emg_traj}A and \ref{fig:hr_emg_traj}B illustrate a single user’s heart rate and EMG trajectories throughout a test session. These figures show that heart rate and raw EMG values are generally higher (in boxed regions) during manual wheelchair use than with \model{}, indicating that manual propulsion requires greater muscular exertion.

\subsection{Muscle Contraction Analysis: Impact of Smart Assistance on Propulsion Efficiency}

\begin{figure}[h]
    \centering
    \includegraphics[width=1\linewidth]{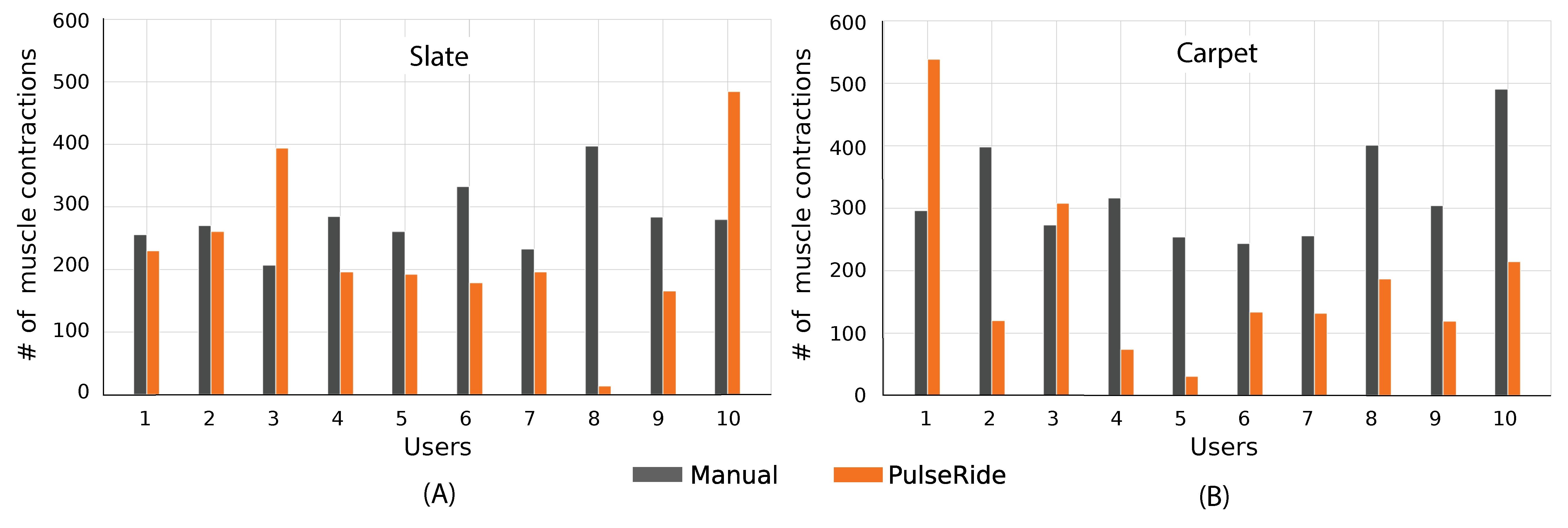}
    \caption{Total number of muscle contractions for each user for manual and PulseRide devices \textbf{A.} on Slate \textbf{B.} on Carpet.}
    \label{fig:push_freq}
    \Description[]{}
    \vspace{-4mm}
\end{figure}

Our muscle contraction analysis, shown in Figure \ref{fig:push_freq} A and B, presents the number of contractions across users and reveals a consistent reduction in muscle effort required to propel \model{} compared to the manual wheelchair condition in both environments (slate and carpet). Across users, \model{} decreased muscle contractions by an average of 41.86\% on slate and 24.94\% on carpet compared to the manual wheelchair. While these reductions indicate that \model{} generally lowers the effort needed for propulsion, some outliers emerged in our results. For example, User 3 and User 10 on slate, and User 1 and User 3 on carpet, did not experience gains in muscle contraction reduction.

\begin{figure}[h!]
    \centering
    \includegraphics[width=\linewidth]{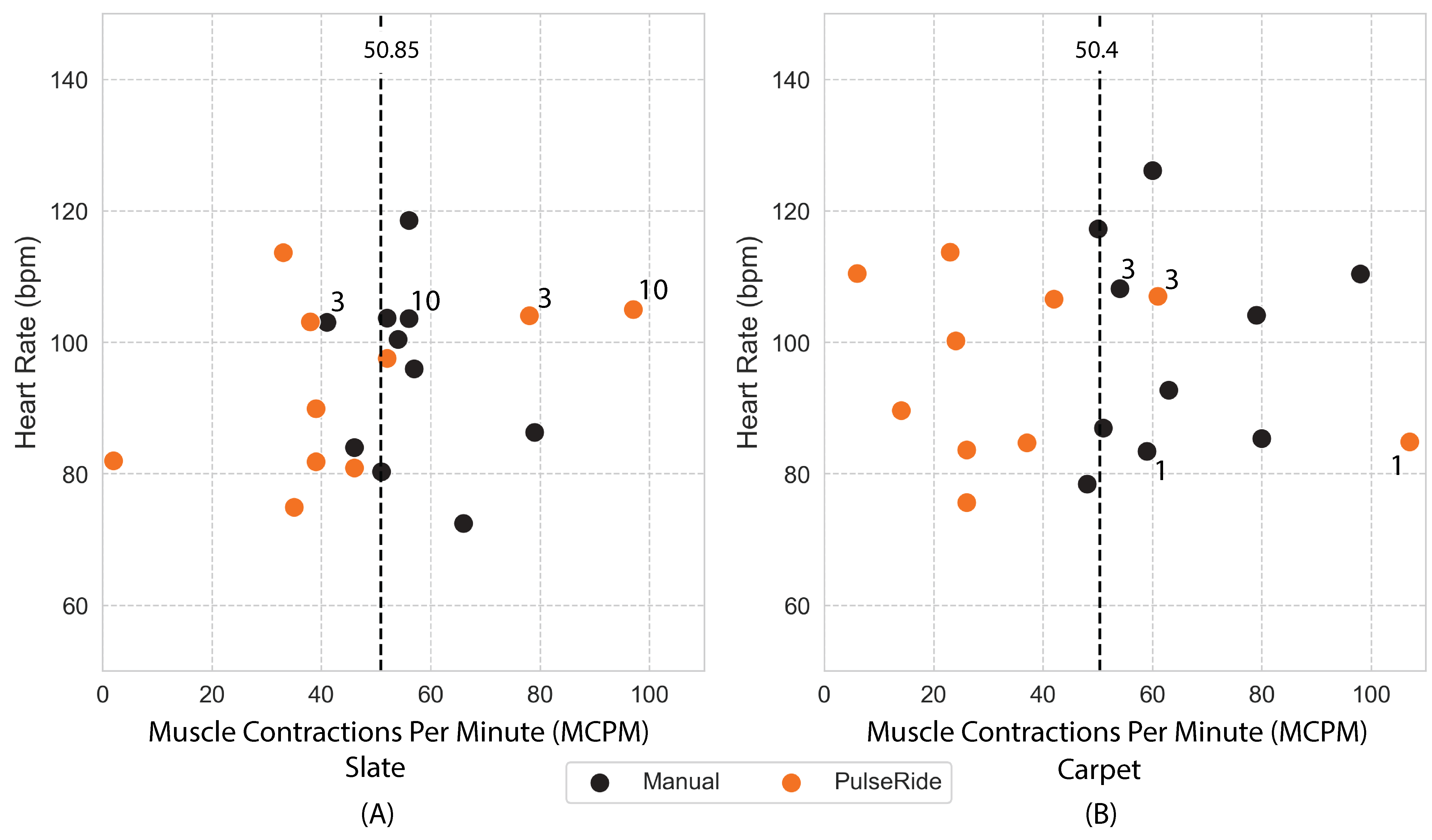}
    \caption{A total number of user muscle contractions per minute for each user for manual and PulseRide wheelchair on the slate surface (shown in \textbf{A}) and carpet surface (shown in \textbf{B})}
    \label{fig:ppm_hr}
    \Description[]{}
\end{figure}

To further understand these outliers, we analyzed muscle contractions per minute relative to heart rate across users. As shown in Figure \ref{fig:ppm_hr} A \& B, User 3 in the slate environment exhibited higher muscle contractions with \model{} than with the manual wheelchair. However, the time this user spent in the moderate heart rate activity zone with \model{} (95.3\%) was similar to the manual condition (94\%). This suggests that, while muscle contractions did not decrease for certain outliers like Users 3, 10, and 1 in the \model{} condition, the intelligent assistance provided by \model{} was deliberate in maintaining a stable heart rate within the moderate activity zone rather than focusing solely on reducing pushes.

Finally, Figure \ref{fig:ppm_hr} A and B highlight distinct clusters for both \model{} and manual conditions, showing that muscle contractions per minute below 50.85\% are associated with \model{} in the slate environment and below 50.4\% in the carpet environment. This clustering further supports the effectiveness of \model{} in reducing muscle effort across the majority of users.


\subsection{Generalization across environments}
\begin{figure}[ht]
    \centering
    \includegraphics[width=1\linewidth]{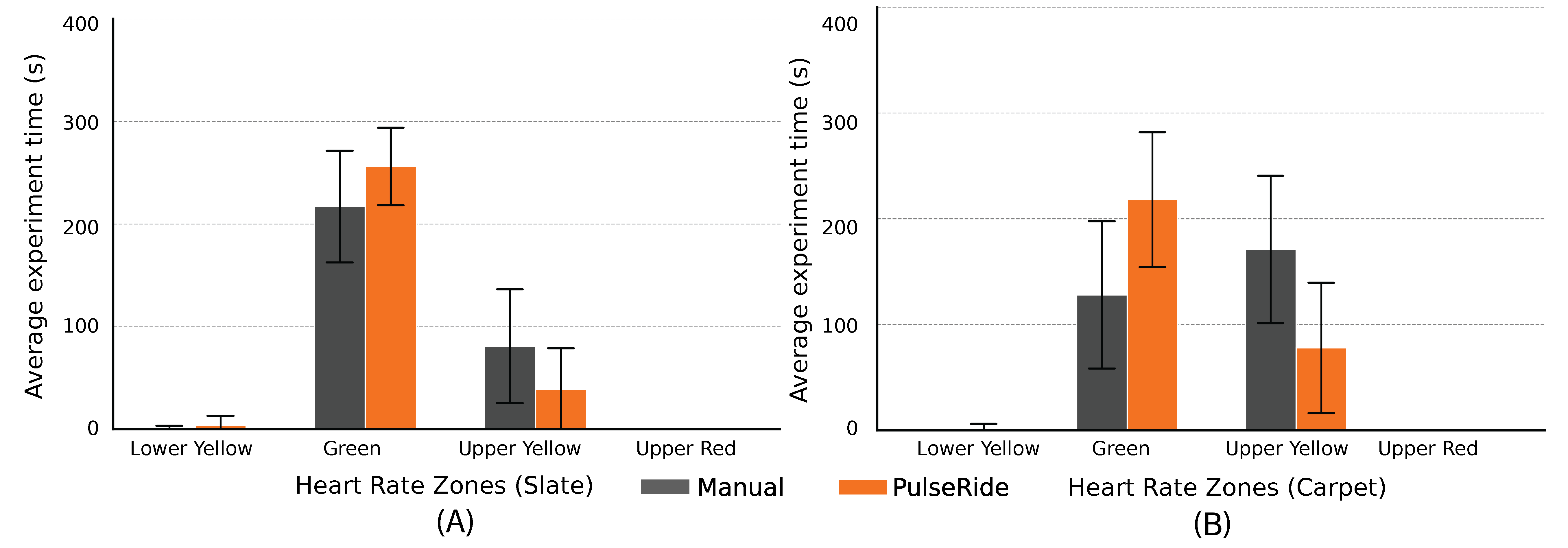}
    \caption{Time spent in heart rate zones during tests (error bars: standard deviation). A. On slate, PulseRide users stayed longer in the moderate activity zone than with manual wheelchairs. B. On carpet, PulseRide also maintained moderate activity better, with manual wheelchairs leading to more intense activity.}
    \label{fig:gen_envs}
    \Description[]{}
    \vspace{-2mm}
\end{figure}

Our results show that across different environments—both slate and carpet—\model{} maintained participants' heart rates within the moderate activity zone. In the slate environment (Figure~\ref{fig:gen_envs}A), participants’ average heart rates were in the Green zone 85.6\% of the time and in the Upper Yellow zone 13.1\% of the time. Similarly, in the carpet environment (Figure~\ref{fig:gen_envs}B), heart rate consistency was preserved, with approximately 73.15\% in the Green zone and 26.2\% in the Upper Yellow zone. This highlights \model{}'s adaptability, as it was trained only in the slate environment yet generalized effectively to the carpet environment.

In contrast, the manual wheelchair showed greater inconsistency across environments. In the slate condition (Figure~\ref{fig:gen_envs}A), participants’ heart rates were in the Green zone 72\% of the time and in the Upper Yellow zone 27\%, following a similar but less efficient trend compared to \model{}. However, on carpet (Figure~\ref{fig:gen_envs}B), the manual wheelchair led to opposite results: participants’ heart rates were in the Upper Yellow zone for approximately 57.2\% of the time, with only 42.8\% in the Green zone. This suggests that participants experienced fatigue more quickly when using a manual wheelchair on carpeted surfaces, making it difficult for them to maintain their heart rate within the moderate zone.



\subsection{PulseRide delays fatigue}

\begin{figure}[h]
    \centering
    \includegraphics[width=1\linewidth]{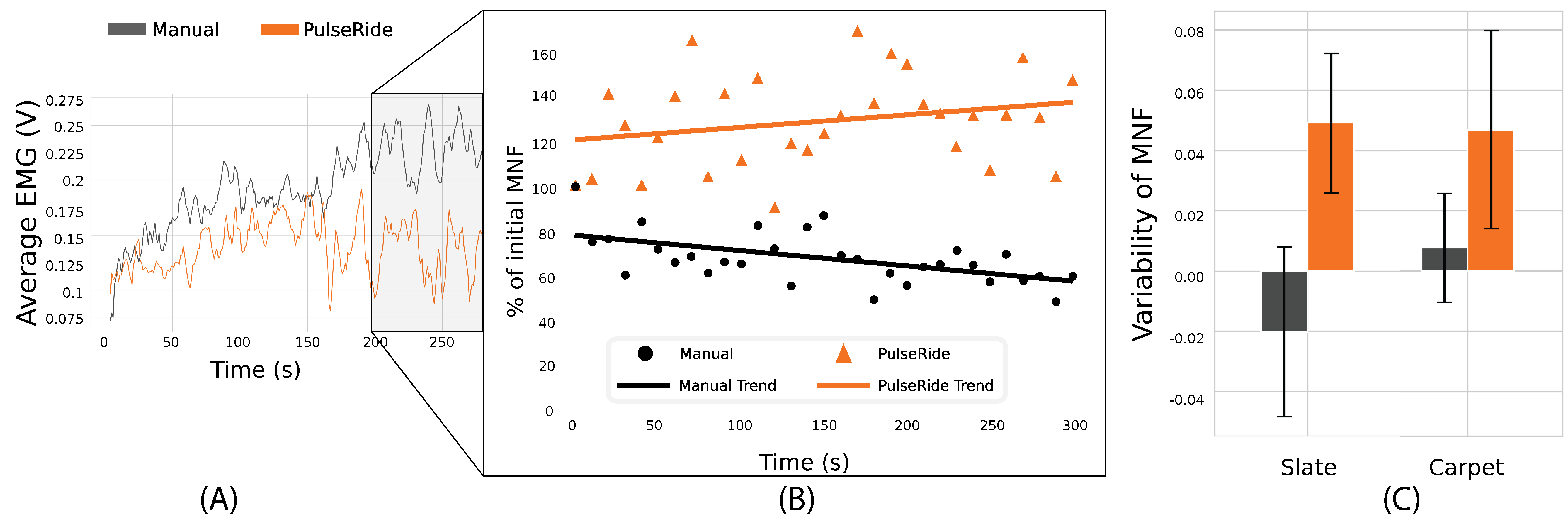}
    \caption{(A) Raw electromyography (EMG) signal with a moving average (T=5) from the triceps brachii of a participant in the carpet environment during PulseRide use and manual wheelchair use. (B) Trend of the mean frequency (MNF) distribution for the participant’s EMG data (10-second window) in the carpet environment for PulseRide and manual wheelchair use. The Y-axis represents a percentage of initial values. A steeper negative slope in the manual wheelchair condition indicates a more rapid onset of fatigue. (C) Variability in mean frequency (MNF) across all participants in both slate and carpet environments for PulseRide and manual wheelchair use. Error bars represent the standard error of the mean.}
    \label{fig:fatigue}
    \Description[]{}
    \vspace{-2mm}
\end{figure}

Our results demonstrate that PulseRide delays fatigue compared to manual wheelchairs. As shown in Figure~\ref{fig:fatigue}A, the rectified average EMG values (moving average with a 5-second window) of a single participant highlight reduced muscle activation during PulseRide use. Figure~\ref{fig:fatigue}B shows the mean frequency (MNF) trend, where the fitted line for PulseRide displays a positive slope, indicating a gradual increase in MNF over time and suggesting reduced fatigue during extended use. Conversely, the MNF trend for manual wheelchairs exhibits a negative slope, indicating increased fatigue toward the end of the session.

Figure~\ref{fig:fatigue}C further illustrates MNF variability across all ten participants in both slate and carpet environments. PulseRide consistently demonstrated more positive MNF variability compared to manual wheelchairs, reflecting less fatigue during prolonged use across all participants. These results underscore PulseRide’s ability to provide adaptive assistance that mitigates fatigue more effectively than manual propulsion.

\section{Discussion}
\label{discussion}


\subsection{Reward plot}
\begin{figure}[h]
    \centering
    \includegraphics[width=0.80\linewidth]{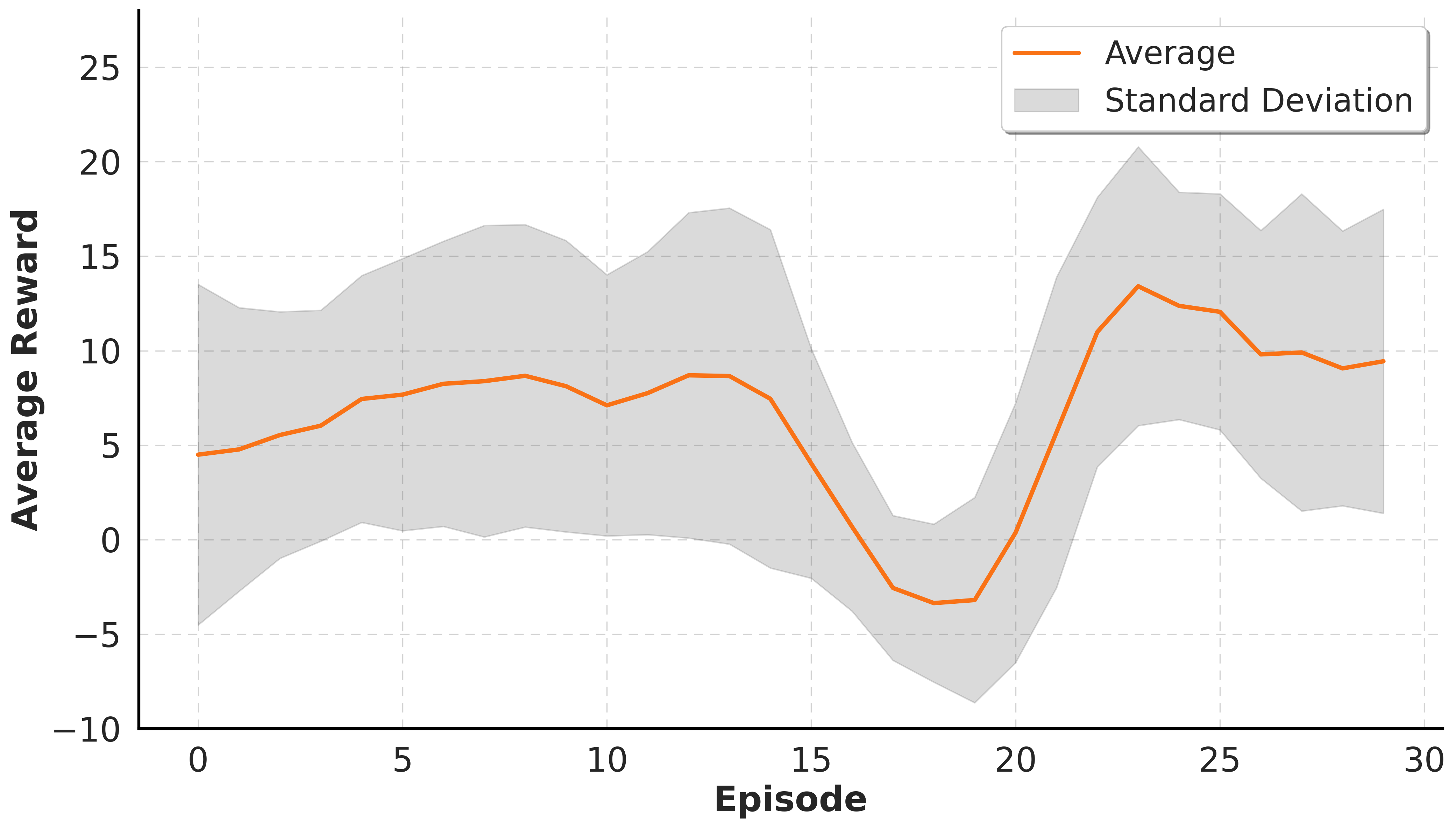}
    \caption{Training reward averaged across all users. The solid orange line represents the average whereas the shaded represents standard deviation.}
    \label{fig:reward}
    \Description[]{}
    \vspace{-2mm}
\end{figure}


Figure \ref{fig:reward} shows the average reward per episode across participants. Initially, the average reward gradually increases up to episode 15 as the agent learns and improves with each episode. To ensure effective policy learning in our human-in-the-loop system, it is crucial that the agent experiences diverse physiological states and velocity ranges during training. This exploration is achieved by prompting the user the interface to follow a structured protocol. At episode 15, participants received a 'Go faster' prompt, followed by a 'Go slower' prompt at episode 25. These prompted phases intentionally drove participants' velocities into transition or extreme zones, resulting in negative rewards. This explains the noticeable dip in rewards from episode 15 to 20. After this high-velocity phase, participants resumed maintaining velocity within their personalized moderate zone, leading to swift increase in reward till episode 25 followed by saturation till episode 30. While the reward trajectory suggests successful policy learning, we restricted the total number of episodes to 30 to keep the training time within 20 minutes. Further investigation is needed to understand potential additional policy improvements with extended training duration.



\subsection{Data Scarcity}
Deep Q-learning algorithms (DQN) typically require substantial environmental interactions to learn an effective policy. However, in this study, training data was limited by participant involvement, with only 600 total interactions available during the training phase. Given that wheelchair propulsion is physically demanding, it was essential to keep the training phase brief while still achieving a well-trained policy. To address this data scarcity, we incorporated domain knowledge through a pre-trained ECG encoder that extracts eight meaningful features per second based on the previous three seconds of electrocardiogram data. This approach enriched the input features available to the primary DQN network. Additionally, we extended the agent's timesteps to two seconds, allowing it to process more data before each update. These adjustments significantly reduced the required training duration while ensuring an optimized model for each participant.




\subsection{Participants without Disabilities}
\label{able_bodied}

Individuals with disabilities typically experience higher exertion and quicker muscle fatigue during wheelchair propulsion due to factors such as reliance on arm exercise, extent of paralysis, reduced sympathetic control, and overall activity levels \cite{haisma2006physical}. While integrating physical activity into daily routines is crucial for cardiovascular fitness and muscle endurance \cite{rimaud2005training}, users are more likely to adhere to exercise intensities of their preference than to adjust based on strict physiological criteria \cite{qi2015wheelchair}. With these considerations in mind, we designed our initial study as a proof of concept with participants without disabilities to evaluate \model{}'s core functionality in a controlled setting. Future studies with wheelchair users are essential to validate these findings in the target population.

\section{Conclusion and Future Work}




In this study, we introduced PulseRide, an adaptive wheelchair system designed to provide personalized assistance based on users' physiological data and movement patterns. Using a Human-in-the-Loop Reinforcement Learning (HITL-RL) approach, PulseRide continuously adjusts push assistance to keep users within their moderate activity heart rate zone, offering a novel solution that bridges the gap between passive and physically taxing mobility options. Our experimental validation demonstrates that PulseRide effectively generalizes across different environments, maintaining consistent performance on both slate and carpet surfaces. Compared to manual wheelchairs, PulseRide successfully reduced muscle contractions and delayed fatigue onset, enabling users to maintain moderate activity levels over extended periods.

Moreover, our analysis shows that PulseRide personalizes assistance to match each user's unique physiological needs, providing tailored support without explicit environment-specific programming. While some outliers were observed, largely due to individual pushing preferences and the absence of braking, PulseRide’s intelligent assistance remained effective in promoting sustained physical engagement without overexertion.

These findings highlight PulseRide’s potential as a viable, adaptive mobility solution that can enhance health outcomes and user comfort, especially for those relying on wheelchairs for daily activities. However, there is still much room for further research, particularly in recruiting individuals with disabilities to provide real-world feedback and performance evaluations. This could broaden the scope of the study, extending its application to rehabilitation, where the system can directly track health and physical improvements over time, automatically suggesting personalized activity plans. While the DQN algorithm used in this study delivered a solid performance, future work should explore more complex algorithms, such as policy gradient \cite{sutton1999policy}, actor-critic \cite{konda1999actor}, and proximal policy optimization methods \cite{schulman2017proximal}, to benchmark performance. Additionally, investigating continuous action spaces for motor thrust tuning could yield even better results.

Overall, this research opens new pathways for the development of assistive technologies driven by physiological data, ultimately promoting physical activity and health for wheelchair users. The potential for future research and system enhancement is vast, and the \model{} system lays the foundation for next-generation personalized assistive devices that are health-conscious.



\bibliographystyle{ACM-Reference-Format}
\bibliography{references}

\appendix

\end{document}